\documentclass{article}

\PassOptionsToPackage{numbers, compress}{natbib}

\usepackage[preprint]{neurips_2025}

\usepackage[utf8]{inputenc}
\usepackage[T1]{fontenc}
\usepackage{hyperref}
\usepackage{url}
\usepackage{booktabs}
\usepackage{amsfonts}
\usepackage{nicefrac}
\usepackage{microtype}
\usepackage{xcolor}
\usepackage[ruled,vlined]{algorithm2e}
\usepackage{amsmath}
\usepackage{graphicx}
\usepackage{array}
\usepackage{pifont}
\usepackage{placeins}
\usepackage{tikz}
\usetikzlibrary{arrows.meta, positioning}

\newcommand{\cmark}{\textcolor{green}{\ding{51}}}
\newcommand{\xmark}{\textcolor{red}{\ding{55}}}

\title{FINESSE: An Agent-Based Simulator and Benchmark Dataset for Multimodal Financial Event Sequences}

\author{%
  Tyler Farnan \quad Benjamin Eng \quad Adam Abate \quad Xirui Hou \\[0.4em]
  \textbf{Rizal Fathony} \quad \textbf{Nam H. Nguyen} \quad \textbf{Senthil Kumar} \\[0.8em]
  Capital One \\[0.8em]
  \small\textbf{Code:} \url{https://github.com/CapitalOne-Research/FINESSE} \\
  \small\textbf{Dataset:} \url{https://huggingface.co/datasets/capitalone/FINESSE}
}

\begin{document}

\maketitle

\begin{abstract}
Machine learning research in financial services is limited by the scarcity of representative open-source datasets. Existing resources are often narrowly focused on a single modality or task and fail to reflect the structured, multimodal, and dynamic nature inherent to many problems in financial services.

In this paper, we introduce \textbf{FINESSE}, a \textbf{F}inancial \textbf{E}vent \textbf{S}equence \textbf{S}imulation \textbf{E}nvironment, an agent-based simulation framework for generating synthetic, structured datasets composed of multiple interdependent event streams. Each stream corresponds to a distinct financial behavior such as transactions, payments, account status changes, and policy interventions, each with unique action spaces, schemas and variable types. These streams are coupled through agents' latent evolving states, enabling the simulation of temporally rich interactions.

We also introduce \textbf{FINESSE-Bench}, a benchmark dataset generated by the simulator, supporting four representative tasks: balance forecasting, transaction fraud detection, missed payment prediction, and next event prediction. We report baseline results using methods from time series forecasting, event sequence modeling, temporal graphs, and temporal point processes. We release the \textbf{FINESSE} framework, including the simulator and dataset to accelerate research on structured, multimodal event sequence modeling challenges in financial services.
\end{abstract}

\section{Introduction}

Synthetic datasets are essential for advancing research on core modeling challenges in the financial services industry \cite{assefa2020generating}. Existing techniques often rely on matching statistical distributions, which raises privacy concerns and limits their applicability. Agent-Based Models (ABMs) offer a principled alternative, enabling the generation of synthetic datasets independent of sensitive data. While ABMs have been applied to financial modeling, their potential for producing structured, multimodal event sequences as benchmark datasets remains underexplored.

Existing financial datasets are often monomodal, designed for narrow use cases, or rely on static account-level attributes. These limitations fail to capture the intricate, evolving, and interdependent event sequences inherent in financial systems. To address this gap, we introduce \textbf{FINESSE}: a \textbf{F}inancial \textbf{E}vent \textbf{S}equence \textbf{S}imulation \textbf{E}nvironment. \textbf{FINESSE} is an agent-based simulation framework designed to generate synthetic benchmark datasets that capture complex and dynamic relationships across financial behaviors, including transactions, payments, account status changes, and policy interventions.

Our work provides a synthetic dataset for evaluating tasks that are representative of underexplored challenges in modeling multimodal event sequence data. Specifically, we contribute:
\begin{itemize}
    \item \textbf{FINESSE-ABM}: A novel, extensible agent-based model for generating long-term, interdependent event sequences with latent state dynamics.
    \item \textbf{FINESSE-Bench}: A synthetic benchmark dataset supporting four representative tasks: spending behavior forecasting, missed payment prediction, transaction fraud detection, and next event prediction.
    \item Initial baseline results for all tasks using state-of-the-art methods, including sequence models, temporal graphs, time-series forecasting, and temporal point processes.
\end{itemize}

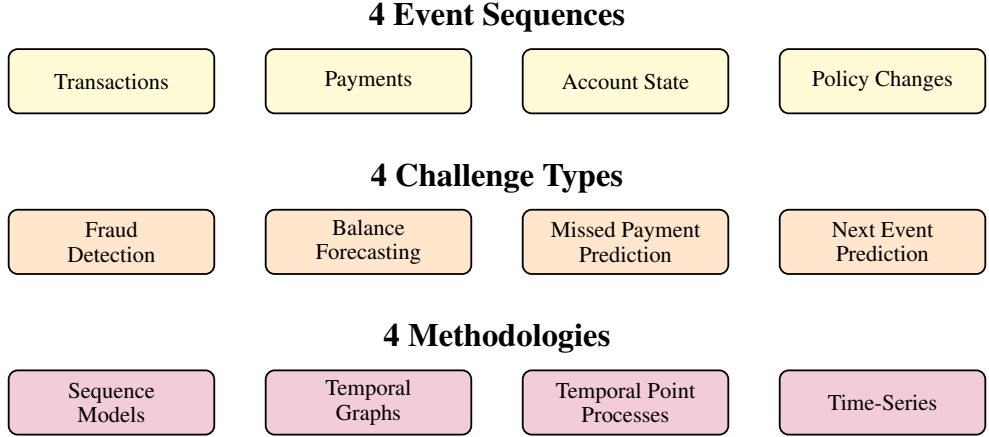
\begin{figure*}[htbp]
  \centering
  \scalebox{0.85}{
  \begin{tikzpicture}[
    box/.style={rounded corners, draw=black, thick, minimum width=3.2cm, minimum height=1cm, align=center},
    seqbox/.style={box, fill=yellow!20},
    chalbox/.style={box, fill=orange!20},
    methbox/.style={box, fill=purple!20},
    node distance=0.8cm
  ]

  \node at (0,6) {\Large \textbf{4 Event Sequences}};
  \node at (0,3.5) {\Large \textbf{4 Challenge Types}};
  \node at (0,1) {\Large \textbf{4 Methodologies}};

  \node[seqbox] (transactions) at (-6,5) {Transactions};
  \node[seqbox] (payments) at (-2,5) {Payments};
  \node[seqbox] (account) at (2,5) {Account State};
  \node[seqbox] (policy) at (6,5) {Policy Changes};

  \node[chalbox] (fraud) at (-6,2.5) {Fraud\\Detection};
  \node[chalbox] (spend) at (-2,2.5) {Balance\\Forecasting};
  \node[chalbox] (missed) at (2,2.5) {Missed Payment\\Prediction};
  \node[chalbox] (next) at (6,2.5) {Next Event\\Prediction};

  \node[methbox] (seq) at (-6, 0) {Sequence\\Models};
  \node[methbox] (graph) at (-2, 0) {Temporal\\Graphs};
  \node[methbox] (tpp) at (2, 0) {Temporal Point\\Processes};
  \node[methbox] (ts) at (6, 0) {Time-Series};

  \end{tikzpicture}
  }
  \caption{FINESSE-Bench High-level Summary}
\end{figure*}
\FloatBarrier

By providing a principled framework and benchmark, \textbf{FINESSE} aims to facilitate research on structured, multimodal event sequence modeling challenges that are underexplored in financial services. The benchmark tasks are designed for research purposes and are not reflective of granular real-world tasks associated with any organization. Additionally, while \textbf{FINESSE} generates synthetic datasets with latent state dynamics, it does not aim to support formal causal inference.

\section{Background}
\subsection{Agent Based Models}
Agent Based models provide a framework for using simple rule sets to generate stochastic, heterogenous, and adaptive decision processes that lead to complex dynamics and emergent phenomena. In 2005, LeBaron’s paper on “Agent-Based Models for Computational Finance” \cite{lebaron2006agent} provides a comprehensive overview of the utility of ABM’s for macroeconomic market dynamics. Similarly, \cite{assenza2015emergent} studies the emergence of macroeconomic fluctuations from capital and credit interactions between large firms and banks. Agent Based methods have also been used for modeling liquidity risk in interbank lending and payment systems \cite{galbiati2011agent} \cite{arciero2008exploring}. 

More recently, ABIDES \cite{dwarakanath2024abides} has introduced reinforcement learning driven agent policies to study household preference to employment at firms and firm resilience to production shocks. The only work directly using ABM’s for credit card customers is focused on modeling customer sensitivity to incentive and promotions \cite{hamill2025agent}. However this work studies public datasets that only contain account-level demographic information, rather than granular event-level patterns of payments and transactions.

Agent-based models (ABMs) have been identified as a promising methodology for generating synthetic data in finance \cite{assefa2020generating}. However, their application has been largely limited to specific use cases, such as Anti-Money Laundering, where known priors of criminal activity are encoded as rules within the simulation \cite{altman2024realistic}. In contrast, our work leverages an ABM-based approach to generate long-term, interdependent event sequences, capturing the dynamics of credit card customer transactions, payments, account states, and policy adjustments. This expands the scope of ABM applications to structured, multimodal event sequence modeling in financial services.

\subsection{Financial Services Datasets}

\subsubsection{Transaction Fraud Detection}  
While there are several open-source financial services datasets, they often have significant limitations. Many focus narrowly on single-modal credit card transaction data for fraud detection, such as datasets providing device-level features \cite{ieeecis2019, fraudecom2019}, or synthetic datasets with basic transaction-level features \cite{sparkov2016, padhi2021tabular}, including PCA-derived features \cite{ccfraud2013}. These datasets have been aggregated into Amazon’s fraud detection benchmark \cite{grover2022fraud}. Other datasets, like \cite{rosbank2024}, propose tasks such as customer attrition and POS-terminal transaction volume but remain limited to fundamental features.

\subsubsection{Loan Default Prediction}  
Loan payment logs, a critical aspect of financial services data, are underrepresented. The Vehicle Loan Default Prediction dataset \cite{vehicleloan2019} provides only account-level credit bureau and demographic features. Similarly, the American Express Kaggle competition dataset \cite{amex2024} offers aggregated statement-level features, including spend, payments, balance, and delinquency. While these datasets are similar in scope, our work provides event-level logs for granular temporal modeling, along with latent state dynamics related to credit card policy adjustments and risky merchants.

\subsubsection{Multimodal Benchmarks}  
Recent efforts, such as FinBench \cite{yin2023finpt}, aggregate single-modal, single-task datasets but lack temporal components, focusing instead on static, account-level features. Similarly, agent-based models (ABMs) like \cite{hamill2025agent} rely on high-level account attributes rather than detailed, interdependent event sequences.

Multimodal datasets like the Berka dataset from the 1999 KDD Cup \cite{pkdd1999} include account info, transactions, payments, loans, and credit cards but lack granular temporal features and policy-driven dynamics. The 2022 Data Fusion Challenge \cite{datafusion2024} combines credit card transactions and clickstream events for stream-matching tasks, whereas our dataset focuses on modeling challenges involving multiple interdependent data streams from the same customer. More recently, the MBD dataset \cite{dzhambulat2024multimodal} introduced multimodal data for campaign analytics and modality matching using transaction logs, geo-tags, and dialogues. While MBD is a significant contribution, our dataset addresses unique challenges in modeling structured, multimodal event sequences, such as balance forecasting, fraud detection, missed payment prediction, and next event prediction. We encourage the community to explore multimodal fusion techniques, like those in MBD, to leverage cross-sequence information for these tasks.

\begin{table}[htbp]
  \caption{A comparison of available data sources in existing multimodal event sequence style benchmark datasets in financial services.}
  \label{tab:data_sources_features}
  \centering
  \begin{tabular}{lcccccc}
    \toprule
    \textbf{Feature} & \textbf{\cite{pkdd1999}} & \textbf{\cite{datafusion2024}} & \textbf{\cite{dzhambulat2024multimodal}} & \textbf{\cite{yin2023finpt}} & \textbf{FINESSE} \\
    \midrule
    Transactions & \cmark & \cmark & \cmark & \xmark & \cmark \\
    Payment & \xmark & \xmark & \xmark & \xmark & \cmark \\
    Credit state & \xmark & \xmark & \xmark & \xmark & \cmark \\
    Policy changes & \xmark & \xmark & \xmark & \xmark & \cmark \\
    Clickstream & \xmark & \cmark & \cmark & \xmark & \xmark \\
    Geotags & \xmark & \xmark & \cmark & \xmark & \xmark \\
    Natural language & \xmark & \xmark & \cmark & \cmark & \xmark \\
    Account info & \cmark & \xmark & \cmark & \cmark & \cmark \\
    \bottomrule
  \end{tabular}
\end{table}
\FloatBarrier

\section{FINESSE-ABM: Simulation Framework for Multimodal Event Sequences}
\textbf{FINESSE-ABM} is an agent-based simulation framework designed to generate structured, multimodal event sequence datasets. It models the interactions between customers, merchants, and banks, capturing the evolving dynamics of customer behavior, merchant risk, and policy interventions. The simulation operates over discrete time steps, with agents transitioning through latent states that influence their actions and interactions. 

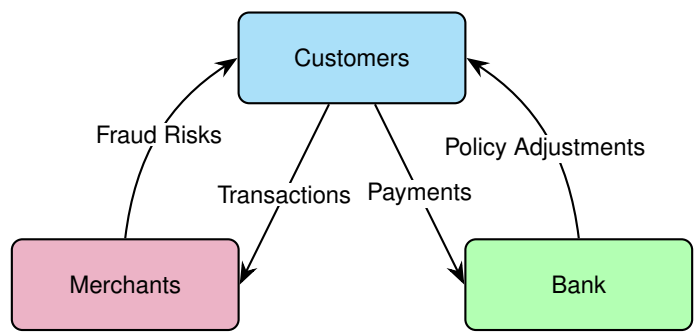
\begin{figure}[htbp]
  \centering
  \label{fig:abm_interaction_overview}
  \begin{tikzpicture}[
    box/.style={
      draw,
      thick,
      rounded corners,
      minimum width=3cm,
      minimum height=1.2cm,
      font=\sffamily\small,
      align=center,
      fill=gray!10
    },
    arrow/.style={-{Stealth[length=3mm]}, thick},
    annot/.style={
      font=\sffamily\footnotesize,
      fill=white,
      inner sep=1pt
    }
  ]

  \node[box, fill=cyan!30] (cust) at (3,1)  {Customers};
  \node[box, fill=purple!30] (mer) at (0,-2) {Merchants};
  \node[box, fill=green!30] (iss) at (6,-2) {Bank};

  \draw[arrow]
    (cust) -- node[annot] {Transactions} (mer.east);

  \draw[arrow]
    (mer.north) to[bend left=25] node[annot] {Fraud Risks} (cust.west);

  \draw[arrow]
    (cust) -- node[annot] {Payments} (iss.west);

  \draw[arrow]
    (iss.north) to[bend right=25] node[annot, below] {Policy Adjustments} (cust.east);

\end{tikzpicture}
  \caption{ABM Interaction Overview}
\end{figure}

In this section, we describe the mechanics of \textbf{FINESSE-ABM}, including its core components illustrated in Figure~\ref{fig:abm_interaction_overview} and the core algorithm, as outlined in Algorithm~\ref{alg:finesse_abm}.

\begin{algorithm}[htbp]
\footnotesize
\setlength{\textfloatsep}{8pt}
\caption{FINESSE-ABM: Agent-Based Simulation for Multimodal Financial Event Sequences}
\label{alg:finesse_abm}
\KwIn{$N_c$: Number of customers, $T$: Number of days, $\mathcal{M}$: Merchants \{MCC, ID, $R_f$, $R_i$\}, $\mathcal{C}$: Customer behaviors, $\text{credit\_limits}[i]$, RCT intervention cohorts.}
\KwOut{$\mathcal{L}_\text{txn}$, $\mathcal{L}_\text{pay}$, $\mathcal{L}_\text{acct}$, $\mathcal{L}_\text{policy}$, $\mathcal{L}_\text{hidden}$.}

\For{$t = 1$ to $T$}{
    \For{$c \in \{1, \dots, N_c\}$}{
        \textbf{Transactions:} If $p_t > \theta_t$, sample $m \sim P(\mathcal{M}; A[c])$, append $\{t, c, m, \text{amt}\}$ to $\mathcal{L}_\text{txn}$, update $R_i[c]$. \;
        \textbf{Payments:} If $p_p > \theta_p$, compute payment from $S[c]$, append $\{t, c, \text{pay}\}$ to $\mathcal{L}_\text{pay}$. \;
        \textbf{Policy Changes:} If $t \mod P = 0$ and $c \notin \text{control}$, sample policy, append $\{t, c, \text{policy}\}$ to $\mathcal{L}_\text{policy}$. \;
        \textbf{Hidden State:} If $p_h < \theta_h$, update $A[c]$ or $S[c]$, append $\{t, c, \text{state}\}$ to $\mathcal{L}_\text{hidden}$. \;
        \textbf{Account State:} Update missed payments and account status, append $\{t, c, \text{acct}\}$ to $\mathcal{L}_\text{acct}$. \;
    }
}
\Return{$\mathcal{L}_\text{txn}, \mathcal{L}_\text{pay}, \mathcal{L}_\text{acct}, \mathcal{L}_\text{policy}, \mathcal{L}_\text{hidden}$}
\end{algorithm}
\FloatBarrier

\subsection{Simulation Setup: Merchants, Customer Behavior, and Bank Policies}

\textbf{Generating Merchants}  
The merchant population is sampled from pre-configured distributions over merchant attributes. These attributes are categorized as follows:  
\begin{itemize}
    \item \textbf{Observable Attributes (recorded in transaction events)}:  
    \begin{itemize}
        \item \texttt{MCC}: Merchant Category Code.  
        \item \texttt{merchant\_id}: Unique identifier for the merchant.  
    \end{itemize}
    \item \textbf{Unobserved Attributes (influence latent state)}:  
    \begin{itemize}
        \item \texttt{popularity}: A measure of how frequently the merchant is selected.  
        \item \texttt{online\_presence}: Likelihood of the merchant being an online vendor.  
        \item \texttt{id\_theft\_risk}: Risk score for identity theft associated with the merchant.  
        \item \texttt{tx\_fraud\_risk}: Risk score for transaction fraud associated with the merchant.  
    \end{itemize}
\end{itemize}

\textbf{Customer Policies}  
Customer behavior is governed by two policies: \textit{Merchant Affinity} and \textit{Payment Strategy}. Each customer is assigned one of $P_{ma}$ Merchant Affinity policies and one of $P_{ps}$ Payment Strategy policies.  
\begin{itemize}
    \item \textbf{Merchant Affinity}: Determines transaction frequency, spending amounts, and online merchant usage within each MCC. Behavioral parameters are sampled from policy-specific distributions, creating distinct customer segments with diverse behaviors.
    \item \textbf{Payment Strategy}: Controls payment frequency and balance paydown proportions, similarly sampled to generate distinguishable yet diverse patterns across segments.
\end{itemize}

\begin{table}[htbp]
  \caption{Heterogeneity parameters for customer agents. Each agent is assigned both a payment strategy type and a merchant affinity type characterized by a parameterized frequency and amount factor distribution.}
  \label{tab:freq}
  \centering
  \begin{tabular}{ccl}
        \toprule
        \textbf{Type} & \textbf{Frequency} & \textbf{Amount} \\
        \midrule
        Type 1 & $F \sim \mathrm{Poisson}(\lambda_1)$ & $A \sim \mathcal{N}(\mu_1,\sigma_1)$ \\
        Type 2 & $F \sim \mathcal{N}(\mu_1,\sigma_1)$ & $A \sim \mathrm{Poisson}(\lambda_1)$ \\
        \vdots  & \vdots & \vdots \\
        Type N & $F \sim \mathrm{Poisson}(\lambda_N)$ & $A \sim \mathcal{N}(\mu_N,\sigma_N)$ \\
        \bottomrule
\end{tabular}
\end{table}
\FloatBarrier

\textbf{Bank Policies}  
To simulate evolving account policies, customers are assigned to either a control or intervention group using randomized control trials (RCTs). Dynamic interventional policies include adjustments to credit limits, interest rates, and minimum payment factors. Static policies such as minimum payment rules and due dates are applied uniformly across all customers.

\FloatBarrier

\subsection{Latent State Dynamics}

\textbf{Customer Behavior Dynamics}  
A key feature of FINESSE-ABM is the evolving hidden state of each customer, which drives the emergence of new behavioral patterns over time. At each time step, customers may probabilistically transition to a new Merchant Affinity or Payment Strategy policy. Bank policy interventions may increase the likelihood of a hidden state transition, depending on the customer's account state.

\textbf{Risky Merchant Interactions}  
The simulation models the interplay between customer behavior and merchant risk profiles. Each merchant is assigned identity theft and fraud risk scores, which influence the likelihood of fraudulent transactions. As customers interact with merchants, their hidden fraud risk score evolves, increasing with repeated transactions at merchants with higher identity theft risk. Over time, the probability of a customer experiencing a fraudulent transaction grows as a function of their evolving fraud risk and the risk profiles of the merchants they engage with.

\section{FINESSE-Bench Dataset Overview}

\begin{table}[htbp]
\centering
\caption{Schemas for each event type in the FINESSE dataset. Each event type contains distinct attributes and is recorded at different frequencies.}
\label{tab:event_schemas}
\begin{tabular}{@{}p{3.5cm}p{8.5cm}p{3cm}@{}}
\toprule
\textbf{Event Type}           & \textbf{Attributes (Name: Type)}                                                       & \textbf{Frequency}       \\ \midrule
\textbf{Transactions} 
& \begin{tabular}[t]{@{}l@{}}
    \texttt{status}: categorical (approved/declined) \\ 
    \texttt{amount}: continuous \\ 
    \texttt{merchant\_category}: categorical \\ 
    \texttt{merchant\_id}: string \\ 
    \texttt{online}: boolean \\ 
    \texttt{fraud}: boolean \\ 
    \texttt{timestamp}: datetime (Y/M/D/H/M/S)
\end{tabular} 
& Per transaction          \\ \midrule
\textbf{Loan Payments}        
& \begin{tabular}[t]{@{}l@{}}
    \texttt{amount}: continuous \\ 
    \texttt{timestamp}: datetime (Y/M/D/H/M/S)
\end{tabular}                                                                                 
& Per payment              \\ \midrule
\textbf{Policy Interventions} 
& \begin{tabular}[t]{@{}l@{}}
    \texttt{intervention\_type}: categorical \\ 
    (credit limit, interest rate, min payment factor) \\ 
    \texttt{timestamp}: datetime (Y/M/D)
\end{tabular} 
& Daily (if applied)       \\ \midrule
\textbf{Account State}        
& \begin{tabular}[t]{@{}l@{}}
    \texttt{current\_balance}: continuous \\ 
    \texttt{utilization\_ratio}: continuous \\ 
    \texttt{interest\_rate}: continuous \\ 
    \texttt{min\_payment\_factor}: continuous \\ 
    \texttt{current\_missed\_payments}: integer
\end{tabular} 
& Daily                   \\ \bottomrule
\end{tabular}
\end{table}
\FloatBarrier

\section{Task Evaluation}

All task evaluations utilized a single A10 GPU, with training times between 12-24 hours.

\subsection{Detecting Fraudulent Transactions}

The particular type of transaction-level fraud we seek to detect occurs via the latent state mechanisms relating identity theft risk to risky merchants, as described in section 3.2. Hence, we cast this transaction-level fraud detection as a binary classification challenge, where the positive label indicates that the transaction is fraudulent; otherwise, it's a normal transaction. We formulate fraudulent transaction detection as edge classification on a temporal graph. 

\subsubsection{Edge Classification on Temporal Graph}

We formulate the transactions history data as a bipartite graph of customers and merchants. The nodes in the graph represent a customer or a merchant, and the edges represent a transaction between a customer and a merchant. In each transaction, the timestamp when the event occurred is provided, forming a temporal graph. Specifically, we formulate the task as an edge classification (fraudulent or not) on a continuous-time dynamic graph (CTDG) \cite{ctdne,dygraph-survey}. 

We run two CTDG-based GNN models as a baseline, the Temporal Graph Networks (TGN) \cite{tgn} and DyRep \cite{dyrep}. We follow the Temporal Graph Benchmarks (TGB) \cite{tgb1,tgb2} implementation of TGN and DyRep with a few modifications on the negative sampling and the memory module update.

In our experiments, we use (70\%, 15\%, 15\%) dataset split (into train, validation, test sets) of the dynamic graph. The split is conducted based on the timestamp, i.e. the training set contains a graph that consists of the first 70\% transaction events. For the performance evaluation, we present both the area under ROC curve and the area under precision-recall curve of the fraud detection task. Table \ref{tbl:ctdg} shows the results for both TGN and DyRep. 

\begin{table}[h]
\centering
\caption{Temporal Graph Baseline Results}
\begin{tabular}{@{}lcc@{}}
\toprule
Model          & Area u. ROC curve          & Area u. Prec-Rec curve         \\
\midrule
DyRep \cite{dyrep}    &  0.7561      & \textbf{0.0435}         \\ \midrule
TGN \cite{tgn}  &  \textbf{0.7607}     & 0.0433        \\
\bottomrule
\end{tabular}
\label{tbl:ctdg}
\end{table}

\subsection{Missed Payment Prediction}

Missed payment prediction is a sequence-level regression task, and is defined as follows. Given historical sequence data for a customer, can we predict the total number of missed payments that will occur N months into the future. A missed payment is determined by whether or not the customer made a minimum payment within their statement due date.

\subsubsection{Baselines}

The baselines used in our sequence classification tasks are the Multi-Layer Perceptron (MLP), Gated Recurrent Unit (GRU), a Transformer \cite{vaswani2017attention}, MAMBA \cite{dao2024transformers}, and CoLES \cite{babaev2022coles}. We train these baselines with default model settings to predict the number of missed payments over a horizon.

\subsubsection{Metrics}

More specifically, the baseline models predict how many missed payments an agent misses within a 3, 6, 9 or 12 month period given their entire transaction and account history. Among the 4 time horizons, we use multiclass ROC-AUC as a measure of classification accuracy.

\subsubsection{Results}

Average multiclass ROC-AUC performance drops overall across almost all baselines as the horizon increases. For the 3 and 6 month horizon, transformers outperform all other models, especially in the 6 month case. At 9 months, the GRU has the best performance, with Transformer coming in second. At 12 months, performance is nearly identical across baseline models, with the GRU performing slightly better than the rest.

\begin{figure}[h]
  \centering
  \includegraphics[width=\linewidth]{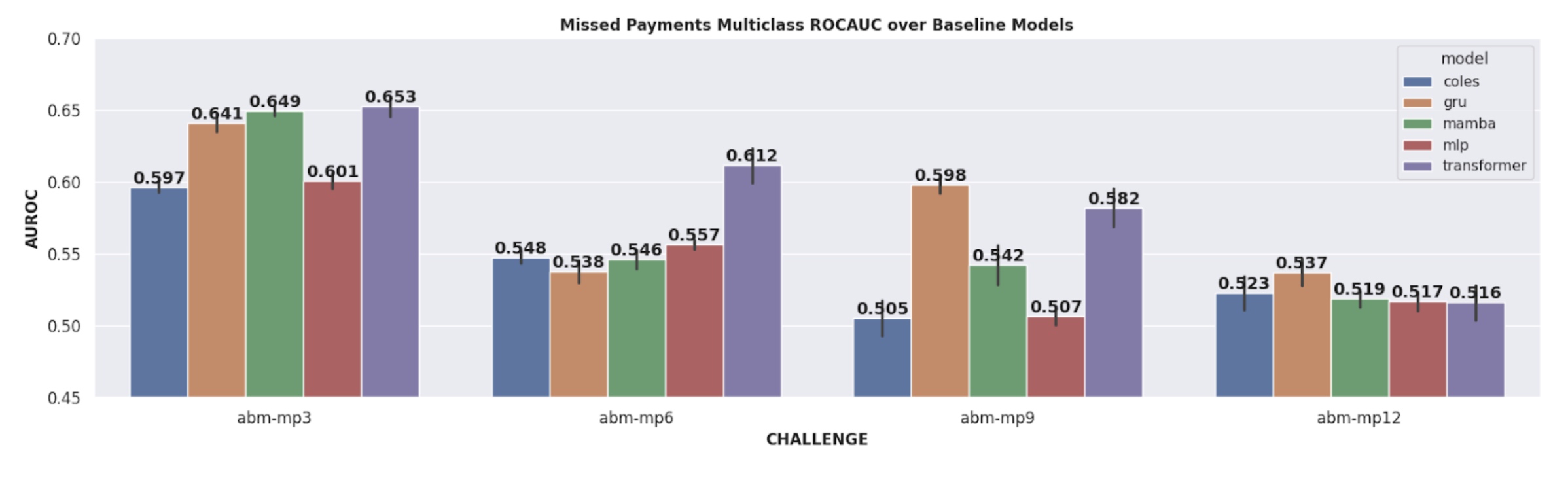}
  \caption{Missed Payment Prediction Baseline Results}
\end{figure}

\subsection{Balance Forecasting}
Balance forecasting is a univariate or multi-variate time-series forecasting problem: given discrete transaction and payment events for a customer, forecast the change in balance patterns over time.

\subsubsection{Experiment Setup}
We use the following baselines: DLinear \cite{zeng2023transformers}, PatchTST \cite{nie2022time}, a Transformer \cite{vaswani2017attention}, and MambaSimple \cite{dao2024transformers}. The inputs to these models are 60-day sequences of past account balances. We train the baselines using default model settings to forecast future values of account balance over 60 days.

\subsubsection{Metrics \& Results}

We evaluate the baselines using MAE, MSE, RMSE, MAPE, and MPSE. PatchTST achieves the best performance in three of five metrics, while the Transformer performs best on MAPE and MPSE. MambaSimple consistently ranks lowest due to the 60-day sequence length not benefiting from its O(L) scaling advantages.

\begin{figure}[h]
  \centering
  \includegraphics[width=\linewidth]{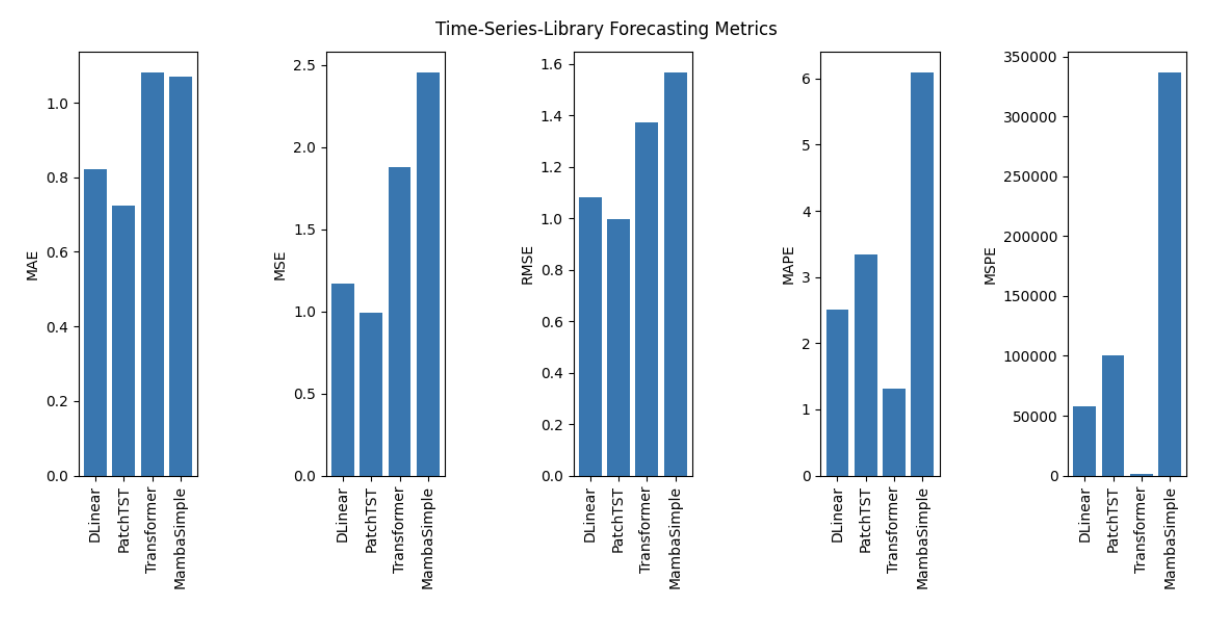}
  \caption{Balance Forecasting Baseline Results}
\end{figure}

\subsection{Next Event Prediction}
Next event prediction is framed as a discrete temporal point process using the EasyTPP package \cite{xue2023easytpp}. We evaluated next payment prediction and next transaction category prediction using Recurrent Marked Temporal Point Process (RMTPP) \cite{du2016recurrent}, Neural Hawkes Process (NHP) \cite{mei2017neural}, Transformer Hawkes Process (THP) \cite{zuo2020transformer}, and Intensity-Free Temporal Point Process (IFTPP) \cite{shchur2019intensity}.

\begin{table}[h]
\centering
\caption{Next-event Type and Time Prediction Baseline Results}
\begin{tabular}{@{}l|cc|cc@{}}
\toprule
& \multicolumn{2}{c|}{Payment}        & \multicolumn{2}{c|}{Transaction}    \\
Model          & Type Error Rate          & RMSE  & Type Error Rate          & RMSE       \\
\midrule
RMTPP \cite{du2016recurrent}    &  0.5245      & 13.7997 & 0.5904 & 0.6606         \\ 
NHP \cite{mei2017neural}  &  0.5241     & 13.5598   & 0.5877 & \textbf{0.6520}    \\ 
THP \cite{zuo2020transformer}  &  0.5080    & 12.5381   & 0.5863 & 0.6576     \\ 
IFTPP \cite{shchur2019intensity}  &  \textbf{0.4922}     & \textbf{12.8061} & \textbf{0.5785} & 1.9132      \\
\bottomrule
\end{tabular}
\label{tbl:tpp}
\end{table}

\section{Conclusion}

We present \textbf{FINESSE}, a novel agent-based simulation framework for generating synthetic, multimodal event sequence datasets with latent state dynamics and evolving complexities. \textbf{FINESSE-ABM} is highly extensible and configurable, enabling researchers to simulate diverse scenarios and explore the relationships between latent states and observable behaviors. Alongside this, we release \textbf{FINESSE-Bench}, a synthetic benchmark dataset designed to evaluate tasks representative of underexplored challenges in financial services, including transaction fraud detection, missed payment prediction, balance forecasting, and next event prediction.

The primary contribution of this work is the establishment of a flexible framework and benchmark that enable research on structured, multimodal event sequence modeling. The baseline results provided serve as an initial reference point. We encourage the research community to build on \textbf{FINESSE} by extending both the simulator and the benchmarking framework.

\bibliographystyle{plainnat}
\bibliography{sample-base}

@article{nie2022time,
  title={A time series is worth 64 words: Long-term forecasting with transformers},
  author={Nie, Yuqi and Nguyen, Nam H and Sinthong, Phanwadee and Kalagnanam, Jayant},
  journal={International Conference on Learning Representations (ICLR)},
  year={2023}
}

@inproceedings{zeng2023transformers,
  title={Are transformers effective for time series forecasting?},
  author={Zeng, Ailing and Chen, Muxi and Zhang, Lei and Xu, Qiang},
  booktitle={Proceedings of the AAAI Conference on Artificial Intelligence},
  volume={37},
  pages={11121--11128},
  year={2023}
}

@inproceedings{du2016recurrent,
  title={Recurrent marked temporal point processes: Embedding event history to vector},
  author={Du, Nan and Dai, Hanjun and Trivedi, Rakshit and Upadhyay, Utkarsh and Gomez-Rodriguez, Manuel and Song, Le},
  booktitle={Proceedings of the 22nd ACM SIGKDD International Conference on Knowledge Discovery and Data Mining},
  pages={1555--1564},
  year={2016}
}

@article{mei2017neural,
  title={The neural hawkes process: A neurally self-modulating multivariate point process},
  author={Mei, Hongyuan and Eisner, Jason M},
  journal={Advances in Neural Information Processing Systems},
  volume={30},
  year={2017}
}

@inproceedings{zuo2020transformer,
  title={Transformer hawkes process},
  author={Zuo, Simiao and Jiang, Haoming and Li, Zichong and Zhao, Tuo and Zha, Hongyuan},
  booktitle={International Conference on Machine Learning},
  pages={11692--11702},
  year={2020},
  organization={PMLR}
}

@article{shchur2019intensity,
  title={Intensity-free learning of temporal point processes},
  author={Shchur, Oleksandr and Bilo{\v{s}}, Marin and G{\"u}nnemann, Stephan},
  journal={arXiv preprint arXiv:1909.12127},
  year={2019}
}

@inproceedings{babaev2022coles,
  title={Coles: Contrastive learning for event sequences with self-supervision},
  author={Babaev, Dmitrii and Ovsov, Nikita and Kireev, Ivan and Ivanova, Maria and Gusev, Gleb and Nazarov, Ivan and Tuzhilin, Alexander},
  booktitle={Proceedings of the 2022 International Conference on Management of Data},
  pages={1190--1199},
  year={2022}
}

@article{vaswani2017attention,
  title={Attention is all you need},
  author={Vaswani, Ashish and Shazeer, Noam and Parmar, Niki and Uszkoreit, Jakob and Jones, Llion and Gomez, Aidan N and Kaiser, {\L}ukasz and Polosukhin, Illia},
  journal={Advances in Neural Information Processing Systems},
  volume={30},
  year={2017}
}

@article{dao2024transformers,
  title={Transformers are SSMs: Generalized models and efficient algorithms through structured state space duality},
  author={Dao, Tri and Gu, Albert},
  journal={arXiv preprint arXiv:2405.21060},
  year={2024}
}

@article{lebaron2006agent,
  title={Agent-based computational finance},
  author={LeBaron, Blake},
  journal={Handbook of Computational Economics},
  volume={2},
  pages={1187--1233},
  year={2006},
  publisher={Elsevier}
}

@article{hamill2025agent,
  title={Agent-based modelling of credit card promotions},
  author={Hamill, Conor Brian and Khraishi, Raad and Gherghel, Simona and Lawrence, Jerrard and Mercuri, Salvatore and Okhrati, Ramin and Cowan, Greig Alan},
  journal={International Journal of Bank Marketing},
  year={2025},
  publisher={Emerald Publishing Limited}
}

@article{assenza2015emergent,
  title={Emergent dynamics of a macroeconomic agent based model with capital and credit},
  author={Assenza, Tiziana and Gatti, Domenico Delli and Grazzini, Jakob},
  journal={Journal of Economic Dynamics and Control},
  volume={50},
  pages={5--28},
  year={2015},
  publisher={Elsevier}
}

@inproceedings{assefa2020generating,
  title={Generating synthetic data in finance: opportunities, challenges and pitfalls},
  author={Assefa, Samuel A and Dervovic, Danial and Mahfouz, Mahmoud and Tillman, Robert E and Reddy, Prashant and Veloso, Manuela},
  booktitle={Proceedings of the First ACM International Conference on AI in Finance},
  pages={1--8},
  year={2020}
}

@article{altman2024realistic,
  title={Realistic synthetic financial transactions for anti-money laundering models},
  author={Altman, Erik and Blanu{\v{s}}a, Jovan and Von Niederh{\"a}usern, Luc and Egressy, B{\'e}ni and Anghel, Andreea and Atasu, Kubilay},
  journal={Advances in Neural Information Processing Systems},
  volume={36},
  year={2024}
}

@article{arciero2008exploring,
  title={Exploring agent-based methods for the analysis of payment systems: A crisis model for StarLogo TNG},
  author={Arciero, Luca and Biancotti, Claudia and d'Aurizio, Leandro and Impenna, Claudio},
  journal={Bank of Italy Temi di Discussione (Working Paper)},
  number={686},
  year={2008}
}

@article{galbiati2011agent,
  title={An agent-based model of payment systems},
  author={Galbiati, Marco and Soram{\"a}ki, Kimmo},
  journal={Journal of Economic Dynamics and Control},
  volume={35},
  number={6},
  pages={859--875},
  year={2011},
  publisher={Elsevier}
}

@article{dwarakanath2024abides,
  title={ABIDES-Economist: Agent-Based Simulation of Economic Systems with Learning Agents},
  author={Dwarakanath, Kshama and Vyetrenko, Svitlana and Tavallali, Peyman and Balch, Tucker},
  journal={arXiv preprint arXiv:2402.09563},
  year={2024}
}

@article{yin2023finpt,
  title={FinPT: Financial Risk Prediction with Profile Tuning on Pretrained Foundation Models},
  author={Yin, Yuwei and Yang, Yazheng and Yang, Jian and Liu, Qi},
  journal={arXiv preprint arXiv:2308.00065},
  year={2023}
}

@article{dzhambulat2024multimodal,
  title={Multimodal Banking Dataset: Understanding Client Needs through Event Sequences},
  author={Dzhambulat, Mollaev and Kostin, Alexander and Maria, Postnova and Karpukhin, Ivan and Kireev, Ivan A and Gusev, Gleb and Savchenko, Andrey},
  journal={arXiv preprint arXiv:2409.17587},
  year={2024}
}

@misc{amex2024,
  author={{American Express}},
  title={American Express - Default Prediction},
  howpublished={\url{https://www.kaggle.com/competitions/amex-default-prediction}},
  year={2022},
  note={Accessed: 2024-06-07}
}

@misc{datafusion2024,
  author={{Data Fusion Contest 2022}},
  title={Data Fusion Contest 2022},
  howpublished={\url{https://ods.ai/tracks/data-fusion-2022-competitions}},
  year={2022},
  note={Accessed: 2024-06-07}
}

@misc{rosbank2024,
  author={{Rosbank ML Competition}},
  title={Rosbank ML Competition},
  howpublished={\url{https://boosters.pro/championship/rosbank1/overview}},
  year={2024},
  note={Accessed: 2024-06-07}
}

@inproceedings{padhi2021tabular,
  title={Tabular transformers for modeling multivariate time series},
  author={Padhi, Inkit and Schiff, Yair and Melnyk, Igor and Rigotti, Mattia and Mroueh, Youssef and Dognin, Pierre and Ross, Jerret and Nair, Ravi and Altman, Erik},
  booktitle={ICASSP 2021-2021 IEEE International Conference on Acoustics, Speech and Signal Processing (ICASSP)},
  pages={3565--3569},
  year={2021},
  organization={IEEE}
}

@article{grover2022fraud,
  title={Fraud Dataset Benchmark and Applications},
  author={Grover, Prince and Xu, Julia and Tittelfitz, Justin and Cheng, Anqi and Li, Zheng and Zablocki, Jakub and Liu, Jianbo and Zhou, Hao},
  journal={arXiv preprint arXiv:2208.14417},
  year={2022}
}

@misc{ieeecis2019,
  title={IEEE-CIS Fraud Detection Dataset},
  author={Vesta Corporation},
  year={2019},
  howpublished={\url{https://www.kaggle.com/c/ieee-fraud-detection}}
}

@misc{ccfraud2013,
  title={Credit Card Fraud Detection Dataset},
  author={European Cardholders},
  year={2013},
  howpublished={\url{https://www.kaggle.com/mlg-ulb/creditcardfraud}}
}

@misc{fraudecom2019,
  title={E-commerce Transaction Dataset for Fraud Detection},
  author={Anonymous},
  year={2019},
  howpublished={\url{https://www.kaggle.com/ntnu-testimon/paysim1}}
}

@misc{sparkov2016,
  title={Sparkov Credit Card Transaction Dataset},
  author={Sparkov Data Generation},
  year={2016},
  howpublished={\url{https://www.kaggle.com/kartik2112/fraud-detection}}
}

@misc{vehicleloan2019,
  title={Vehicle Loan Default Prediction Dataset},
  author={Anonymous},
  year={2019},
  howpublished={\url{https://www.kaggle.com/zaurbegiev/my-dataset}}
}

@misc{pkdd1999,
  title={Workshop Notes on Discovery Challenge PKDD'99},
  author={Berka, Petr},
  year={1999},
  howpublished={\url{https://sorry.vse.cz/~berka/challenge/pkdd1999/chall.htm}}
}

@article{xue2023easytpp,
  title={Easytpp: Towards open benchmarking the temporal point processes},
  author={Xue, Siqiao and Shi, Xiaoming and Chu, Zhixuan and Wang, Yan and Zhou, Fan and Hao, Hongyan and Jiang, Caigao and Pan, Chen and Xu, Yi and Zhang, James Y and others},
  journal={arXiv preprint arXiv:2307.08097},
  year={2023}
}

@inproceedings{dyrep,
  title={Dyrep: Learning representations over dynamic graphs},
  author={Trivedi, Rakshit and Farajtabar, Mehrdad and Biswal, Prasenjeet and Zha, Hongyuan},
  booktitle={International Conference on Learning Representations (ICLR)},
  year={2019}
}

@article{tgn,
  title={Temporal graph networks for deep learning on dynamic graphs},
  author={Rossi, Emanuele and Chamberlain, Ben and Frasca, Fabrizio and Eynard, Davide and Monti, Federico and Bronstein, Michael},
  journal={arXiv preprint arXiv:2006.10637},
  year={2020}
}

@inproceedings{ctdne,
  title={Continuous-time dynamic network embeddings},
  author={Nguyen, Giang Hoang and Lee, John Boaz and Rossi, Ryan A and Ahmed, Nesreen K and Koh, Eunyee and Kim, Sungchul},
  booktitle={Companion Proceedings of the The Web Conference 2018},
  pages={969--976},
  year={2018}
}

@article{dygraph-survey,
  title={Representation learning for dynamic graphs: A survey},
  author={Kazemi, Seyed Mehran and Goel, Rishab and Jain, Kshitij and Kobyzev, Ivan and Sethi, Akshay and Forsyth, Peter and Poupart, Pascal},
  journal={Journal of Machine Learning Research},
  volume={21},
  number={70},
  pages={1--73},
  year={2020}
}

@article{tgb1,
  title={Temporal graph benchmark for machine learning on temporal graphs},
  author={Huang, Shenyang and Poursafaei, Farimah and Danovitch, Jacob and Fey, Matthias and Hu, Weihua and Rossi, Emanuele and Leskovec, Jure and Bronstein, Michael and Rabusseau, Guillaume and Rabbany, Reihaneh},
  journal={Advances in Neural Information Processing Systems},
  volume={36},
  year={2024}
}

@article{tgb2,
  title={TGB 2.0: A Benchmark for Learning on Temporal Knowledge Graphs and Heterogeneous Graphs},
  author={Gastinger, Julia and Huang, Shenyang and Galkin, Mikhail and Loghmani, Erfan and Parviz, Ali and Poursafaei, Farimah and Danovitch, Jacob and Rossi, Emanuele and Koutis, Ioannis and Stuckenschmidt, Heiner and others},
  journal={arXiv preprint arXiv:2406.09639},
  year={2024}
}

\end{document}